\documentclass[11pt]{article}
\usepackage{graphicx} 

\title{ExpressivityBench: Can LLMs Communicate Implicitly?}

\usepackage{authblk}

\usepackage{changepage,threeparttable}

\usepackage{tcolorbox}
\usepackage{float}
\usepackage{subfloat}
\usepackage{subfig}

\usepackage{booktabs}
\usepackage{amsmath}
\usepackage{multirow}
\usepackage{multicol}
\usepackage[round]{natbib}
\usepackage[final]{acl}

\usepackage{times}
\usepackage{latexsym}
\usepackage{amsfonts}
\usepackage{algorithm2e}
\usepackage[T1]{fontenc}

\usepackage[utf8]{inputenc}

\usepackage{microtype}

\usepackage{inconsolata}

\author[$\heartsuit$]{Joshua Tint\thanks{This work was completed prior to affiliation with Amazon.}}
\author[$\spadesuit$]{Som Sagar}
\author[$\spadesuit$]{Aditya Taparia}
\author[$\spadesuit$]{Kelly Raines}
\author[$\spadesuit$]{\authorcr Bimsara Pathiraja}
\author[$\spadesuit$]{Caleb Liu}
\author[$\spadesuit$]{Ransalu Senanayake}

\affil[$\heartsuit$]{Amazon}  
\affil[$\spadesuit$]{Arizona State University}  

\affil[ ]{\texttt{\{jrtint, ssagar6, ataparia, kraine5, bpathir1, calebliu, ransalu\}@asu.edu}}

\begin{document}

\maketitle

\begin{abstract}
    Human communication is often implicit, conveying tone, identity, and intent beyond literal meanings. While large language models have achieved strong performance on explicit tasks such as summarization and reasoning, their capacity for expressivity, or implicit communication, remains underexplored. We introduce \textbf{ExpressivityBench}, a framework for evaluating the expressivity of LLMs using information-theoretic communication models. Our approach quantifies how well LLM-generated text communicates target properties without explicit mention, across nine tasks spanning emotion, identity, and tone. To enable scalable and reproducible evaluation, we employ LLM-based graders validated against human judgments. Our results reveal that while models are adept at expressing affective content, they struggle with sociolinguistic signals, lagging behind human baselines. This study provides a necessary step to evaluate human-like implicit communication, with implications for applications such as education, mental health support, and socially-aware dialogue systems. We provide code and data for our benchmark alongside our paper. 

\end{abstract}

\section{Introduction}

Much of human communication is implicit. The phrasing and tone of a message can convey a number of topics beyond its literal meaning: shaping tone, signaling social identity, or subtly guiding interpretation \cite{knepper2017implicit}. When a doctor tells a patient, ``You should make some lifestyle changes,'' versus, ``There are some adjustments that could really benefit your health,'' the explicit message remains the same, but the implied tone and emotional impact differ. This ability to communicate implicitly is a core aspect of natural language use, yet it remains underexplored in Large Language Models (LLMs).

LLMs \cite{OpenAI2023GPT4TR, touvron2023llama} have transformed fields reliant on human communication, including education \cite{OpenAI2023GPT4TR}, customer support \cite{radford_language_2019}, legal services \cite{chern2024can}, and healthcare \cite{bubeck2023sparks}. Chatbots in these fields interact with humans in potentially sensitive and stressful situations, demanding user trust, which is most freely given if the LLM behaves in a human-like way \cite{trustllm, ding2025citations}. As models grow in size and capability, their performance is typically evaluated on explicit tasks like translation, summarization, and question-answering \cite{devlin2018bert, brown2020language}, or on basic linguistic competence or reasoning tasks \cite{davies2023competence, ziyu2023through, hao2024llm}. However, true human-like communication requires more than factual accuracy—it requires \emph{expressivity}, the ability to convey implicit information \cite{ModernLinguisticExpressiveness}.

Expressivity influences not only tone but also social meaning. A model’s word choice (e.g. formal, colloquial, or slang) can implicitly communicate a speaker’s regional background, education level, or identity \cite{green2016slang}. Recognizing this, developers have begun incorporating controls for persona-based expressivity; OpenAI, for example, introduced a feature into its ChatGPT models in Janurary 2025 allowing users to adjust ``personalities'' with settings like ``chatty'' or ``Gen-Z'' \cite{ferguson2025chatgpt}. Despite these advances, expressivity remains poorly understood in LLMs, in particular whether models can accurately convey these implicit tones and signals. Studying how models communicate implicitly is critical for improving their ability to generate ``human-like'' responses—enhancing trust, usability, and alignment with human expectations \cite{trustllm}.

\begin{figure*}[t]
    \centering
        \caption{ExpressivityBench tests LLMs on their ability to implicitly express information using an information theoretic channel method, measuring a generator's ability to faithfully convey implicit signal to a grader.}
    \includegraphics[width=0.9\textwidth]{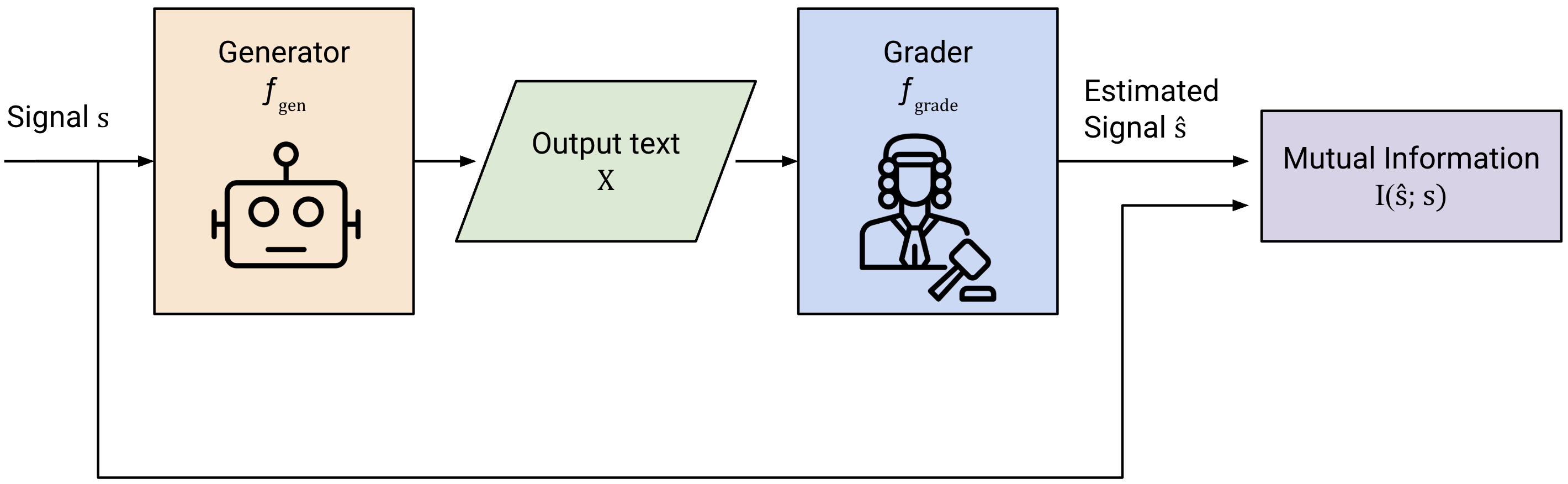}
    \label{fig:graybox}
\end{figure*}

This study aims to quantitatively measure the expressivity of state-of-the-art LLMs. To this end, we focus on the following research question: \textbf{How expressive are LLMs when compared to humans? }

In order to answer this question, we present \emph{ExpressivityBench}, a framework to evaluate expressivity in LLMs. We use information theoretic communication models to measure quantitatively the level of expressivity in LLM generated text. Our method employs a \emph{grader} to evaluate \cite{benedetto2013readers} outputs generated by target LLMs on nine tasks. For reproducibility and scalability~\cite{lee2023rlaif, bai2024benchmarking, chern2024can}, we use LLMs as evaluators to determine how strongly implicit signals are expressed.  We perform our task and grader selection through a human study and establish human-written text as a baseline, in order to understand how LLMs compare to humans. We find that LLMs perform strongly on commonplace expressivity tasks like emotional text generation, but perform much poorer on more complex identity or sociolinguistic-related tasks.

\section{Related Work}

\subsection{Defining Expressivity}

In this section, we explore the literature for definitions of expressivity related to natural language processing in order to formalize our own definition of expressivity. Study of expressivity in NLP is typically limited to emotions, as studied in affective computing devices~\cite{pyreddy2025emoxpt, hu2024recent}. This includes recognizing and emulating emotions from written language~\cite{plaza-emotion-analysis}. However, this limited focus does not capture other aspects that we use in our day-to-day communication, such as tone, metaphor, stylistic variation, and implicit social cues. We build a definition inclusive of this variation.

Thus, we adapt a definition from linguistics, to term ``expressivity'' as the state of communicating information implicitly: \emph{showing, not telling}~\cite{EssentialsOfLinguistics}. To further clarify, Yus distinguishes implicit from explicit communication by noting that implicit meaning must be inferred using context and pragmatics, whereas explicit meaning is directly represented in semantics \cite{yus1999misunderstandings}. For example, ``cheap'' and ``affordable'' share a literal meaning but differ in connotation, while ``greetings'' conveys more formality than ``hello.'' Since LLMs may struggle with contextual interpretation, studying expressivity helps reveal their linguistic limitations \cite{zhu-etal-2024-large}. Other factors are often implicitly inferred as they are read, including sociolinguistic traits \cite{stecker2001expressiveness}. A writer's intended audience is often implicitly communicated as well, for instance one would speak differently to children rather than to adults \cite{gutt1996implicit}. From the literature, we note four broad categories of expressivity: communicating the speaker's intention (e.g. intended genre), speaker's audience (e.g. level of formality), speaker's identity (e.g. age and gender) and speaker's current state (e.g. current emotion) \cite{deckert2023various, choi2020ten}.

\subsection{Evaluating Large Language Models}
Existing benchmarks for LLMs evaluate their performance across diverse tasks, including mathematics \cite{collins2023evaluating}, logical reasoning \cite{parmar2024logicbench}, and education \cite{dai2023can}. These benchmarks typically rely on either automated evaluation using external LLMs \cite{lin2023llmeval,verga2024replacing} or human feedback, as seen in Chatbot Arena \cite{chiang2024chatbot}. Automated evaluation has gained traction due to its scalability and efficiency \cite{surveyevalllm}, and AI feedback has been proposed as a scalable alternative to RLHF \cite{tunstall2023zephyr,lee2023rlaif}. While automated evaluation can induce biases, it also improves reproducibility \cite{sharma2024critical}. In order to mitigate weaknesses of automated evaluation, \citet{laskar-etal-2024-systematic} recommends reporting detailed model versions, validating grader models on gold labels, and sharing evaluation data, which we practice for this study.

Evaluating language models' understanding of pragmatic communication and non-literal language remains challenging. Studies on figurative language comprehension have explored scene modeling for interpreting metaphors and idioms, and related tasks like dereferencing metonomies \cite{gu-etal-2022-just, chakrabarty-etal-2022-flute, sravanthi2024pubpragmaticsunderstandingbenchmark}. Benchmarks such as SocKET assess social communication aspects like humor and sarcasm \cite{choi-etal-2023-llms}, while DailyDialog evaluates conversational abilities \cite{li-etal-2017-dailydialog}. Research has highlighted difficulties in teaching models empathy due to data limitations \cite{rashkin-etal-2019-towards} and their tendency to favor literal interpretations over pragmatic understanding \cite{hu-etal-2023-fine}. However, most pragmatics-oriented benchmarks are largely discriminative, focusing on LLMs' ability to discern pragmatic information \cite{ma2025pragmaticseralargelanguage}.

\subsection{Constrained Generation Tasks}

Constrained generation tasks assess models' ability to produce text which satisfies natural-language conditions, for example generating text which must implicitly communicate a signal. Recent work has introduced modular frameworks like COLLIE \citep{yao2023colliesystematicconstructionconstrained} to define compositional constraints across multiple linguistic levels, enabling more interpretable and flexible experimentation. Nevertheless, integrating complex constraints into LLMs remains a substantial challenge. Several studies highlight limitations in the ability of autoregressive models to maintain multiple constraints over longer generations \citep{chen2024evaluatingunderstandingimprovingconstrained, garbacea2025constrainedneurallanguagegeneration}, and the difficulty of balancing fluency and constraint satisfaction \citep{liu2024we}. These challenges are especially salient in expressive generation tasks, where models must not only satisfy abstract stylistic or sociolinguistic constraints but do so in a manner that appears natural and contextually appropriate to human readers.

\section{ExpressivityBench}

\subsection{Overview}

We take an information-theoretic approach to understanding expressivity, adopting a method derived from the ``encoder-channel-decoder'' model. Classically, these setups involve an ``encoder'' mechanism that communicates a signal through a channel, which must then be received and interpreted by a ``decoder.'' The channel's capacity to communicate information can be derived from the similarity of the final transcribed message to its original signal. We adapt this to measure how well a model can communicate an implicit signal in a piece of natural language text. Formally, the goal of an expressivity experiment is to measure whether a \emph{generator}, $f_{gen}(\cdot)$ can produce a piece of text in the domain $d$, implicitly containing a signal $s$ strongly enough that it can be discerned by a \emph{grader} $f_{grade}(\cdot)$ from a set of other signals $\mathcal{S}$. 

To start, our benchmark has a domain $d$, which is a string representing the type of text that the model must generate (e.g. ``song,'' or ``recipe'') and a signal category $\mathcal{S}$, a set which contains all possible signals which might be communicated in the experiment. A particular signal $s \in \mathcal{S}$ is selected, and the generator $f_{gen}(\cdot)$ is used to generate text with the following prompt $W$:

\begin{figure}[h]
    \centering
    \begin{tcolorbox}[colback=gray!10, colframe=gray!50!black]
    ``Write a $\langle d \rangle$ which conveys $\langle s \rangle$. Do not explicitly mention $\langle s \rangle$ in your response. Do not convey any of the following signals: $\langle  \mathcal{S} \setminus s \rangle$''
    \end{tcolorbox}
    \label{fig:graybox}
\end{figure}

The result, $X=f_{gen}(W)$, is the text that must be evaluated by the grader. To avoid unintentionally leaking $s$ in the response, if $X$ contains the signal $s$, then the response is regenerated. The blind grader, $f_{grade}(\cdot)$ is then prompted to select, of all the signals in $\mathcal{S}$, which one is present in $X$, using the following prompt:

\begin{figure}[h]
    \centering
    \begin{tcolorbox}[colback=gray!10, colframe=gray!50!black]
        ``Which of the following is conveyed in the text: $\langle \mathcal{S} \rangle$?
        Here is the text: $\langle X \rangle$
        Now respond ONLY with one of the terms from this list (do not say none) and answer with no preamble: $\langle \mathcal{S} \rangle$?''
    \end{tcolorbox}

    \label{fig:graybox2}
\end{figure}

Its selection is the estimated signal $\hat{s}$. We repeat these experiments for a fixed number of times over all signals in $\mathcal{S}$ in order to create an induced probability distribution $p(s|\hat{s})$, i.e., the probability that, given the grader guessed some estimated signal $\hat{s}$, the intended signal was actually $s$. In order to measure how expressive the generator $f_{gen}(\cdot)$ is, we take the mutual information over this distribution $I(s;\hat{s})$ (A worked example of mutual information is in Appendix \ref{app:mi-worked-example}). This represents the number of bits of information that the signal $s$ communicates towards the grader's guess, or equivalently, the number of bits of information that the grader's guess tells us about which signal was originally communicated. A high mutual information score indicates that the generator can communicate implicit signals with strong fidelity. We also compute the normalized mutual information, $N=I(s; \hat{s}) / H(s)$, where $H(s)$ is the entropy of the ground truth labels. This normalized value represents the proportion of the total possible information that is actually attributable to the signal. A score of 1 would indicate perfect communication fidelity, while a score of 0 would indicate no signal transfer.

\subsection{Human Study}

\begin{adjustwidth}{-2.5 cm}{-2.5 cm}\centering\begin{table*}[!htb]

\caption{List of tasks used in the human study}\label{tab:categories}

\begin{tabular}{p{3cm}p{6cm}p{2cm}lp{1cm}p{1cm}}\toprule
Signal category &Signals &Text domain &\# Samples &Source \\\midrule
emotion &sad, love, peace, joy, courage, surprise, hate &poem &450 & [\citenum{sreeja2019perc}] \\
register &formal, casual &text message &500 & [\citenum{pavlick2016empirical}] \\
genre &science fiction, thriller, romance, humor, fantasy &short story excerpt &300 & Goodreads \\
MPAA rating &PG, G, PG-13 &short story excerpt &500 & [\citenum{crossley2021commonlit}] \\
tone &cautionary, witty, apathetic, apologetic &line of dialogue &554 & [\citenum{atif2023tone}] \\
skill level &high writing quality, low writing quality &argumentative essay &500 & [\citenum{crossley5129353large}] \\
political slant &left, center, right &news headline & 500 & [\citenum{haak2023qbias}] \\
age &author under 20 years old, author between 20 and 30 years old, author above 30 years old &blog post &500 & [\citenum{schler2006blogging}] \\
gender &author is male, author is female &blog post &500 & [\citenum{schler2006blogging}] \\
hobby (not used in benchmark) &author likes to cook, author likes to volunteer, author likes to hike, author likes to travel &line of dialogue &505 & [\citenum{zhang2018personalizingdialogueagentsi}] \\
profession (not used in benchmark) &author is an engineer, author is a manager, author is a lawyer, author is a musician &line of dialogue &504 & [\citenum{zhang2018personalizingdialogueagentsi}]\\
\bottomrule
\end{tabular}
\end{table*}\end{adjustwidth}

In order to extend this experiment structure into a benchmark, we needed to select a set of tasks and a grader to evaluate with. In the interest of modeling human expressivity as closely as possible, we rooted these decisions in a human study that evaluated the ability of humans to recognize expressivity in a variety of tasks. This serves two purposes: understanding which types of expressive signals can be discerned by real humans, and understanding how closely various automated graders match real human graders' decisions. 

To select an initial list of tasks for this human study, we sought out datasets of text annotated with signals. The criteria for inclusion were:

\begin{enumerate}
    \item Text in the dataset had to be written by humans
    \item The signal annotation had to be performed by humans
    \item The type of signal used had to correspond to some dimension of expressivity: i.e., convey implicit information about the speakers' intention, audience, background or state. Many datasets fell under multiple of these categories.
\end{enumerate}

We selected 10 datasets covering a diverse range of text domains and types of expressivity. Additionally, we were strongly motivated to include fiction genre as an expressivity category due to its prevalence in creative generation \cite{li2025dr}, but we were unable to find a dataset that matched our criteria, so we collected one from the Goodreads website. Details about this dataset collection can be found in Appendix \ref{goodreads}.

In the interest of keeping our tasks tractable and ensuring high automated grader quality, we limited the number of signals in each task. For tasks that had large numbers of signals annotated in the dataset (hobby, profession, genre, emotions, and tones), we used an algorithm described in Appendix \ref{winnowing} to extract a set of maximally distinct labels. For instance, the original Tone Analysis dataset had over 20 labels; we extracted 4. Some datasets had well over 5,000 entries (register, tones, skill level, political slant, age, and gender), so these had random samples of 500 texts taken. 

For \textit{emotion}, we drew from the PERC dataset, which contains poems labeled with one of seven emotions. There are many datasets of emotion-tagged text, but we chose this poetry-based one because poetry is typically thought of as a very expressive domain, and because it would add more diversity to the domains used in the benchmark \cite{sreeja2019perc}. The \textit{register} task was constructed using a 500-sample subset of text messages from the Pavlick Formality Scores dataset, which contains messages annotated as ``formal'' or ``casual'' \cite{pavlick2016empirical}. Formality is particularly useful for analyzing stylistic nuance \cite{ yus1999misunderstandings}. For \textit{genre}, we identified five literary categories from Goodreads, leveraging a random sample of 300 short story excerpts to assess narrative inference (see Appendix \ref{goodreads}). To examine expressivity with respect to intended audience, we included an \textit{MPAA} category task using 500 excerpts from the CLEAR Corpus, which consists of short stories and narrative texts written for U.S. schoolchildren of various ages \cite{crossley2021commonlit}. Each excerpt is labeled with a rating analogous to the Motion Picture Association of America (MPAA) system: G (appropriate general audiences), PG (parental guidance suggested), or PG-13 (some material may be inappropriate for children under 13).  For \textit{tone}, we utilized 554 lines of dialogue labeled with pragmatic attitudes like cautionary, witty, or apologetic, sourced from the Tone Analysis dataset \cite{atif2023tone}. In contrast to emotions, tone is a more intentional reflection of an author's attitude towards a subject \cite{greene2023teaching}. The \textit{skill level} task was framed around writing quality, with samples from argumentative essays in the ASAP-2.0 dataset which were labeled according to grades received in an essay competition \cite{crossley5129353large}. In the original dataset, essays were scored from 1-6. We assigned natural-language labels ``low quality writing'' to essays with a score $\leq$ 2 and ``high quality writing'' to essays with a score $\geq$ 4. Essays with a score of 3 exactly were discarded. This categorization scheme corresponds as closely as possible with the top third and bottom third of essays. \textit{Political slant} was assessed on 500 news headlines from the QBias dataset, which were labeled as leaning left, center, or right \cite{haak2023qbias}. Our final four categories focus on testing the ability of models to infer personal information embedded in narrative voice. For \textit{age} and \textit{gender}, we used 500 blog posts each from the Blog Authorship Corpus, labeled according to three age brackets (under 20, 20–30, over 30) and binary gender (male, female), respectively \cite{schler2006blogging}. While many text datasets are annotated by gender and age, we chose this blogging-based corpus because blogging is a more personal medium, and therefore personal traits may be better reflected by it. Finally, we used the PersonaChat dataset for two classification tasks; this dataset consists of human-human conversations tagged by details of each interlocutor's persona. From these we created a \textit{hobby} task, based on speakers' preferences like cooking or traveling, and \textit{profession} such as engineer or musician \cite{zhang2018personalizingdialogueagentsi}. Across all categories, we filtered out samples which contained the name of the signal in the text, matching the regeneration criteria for automated generation. A summary of the tasks can be found in Table \ref{tab:categories}. We provide all the human text samples as part of our benchmark along with our paper.

For our human study, each participant was presented with two random text samples from each of the 11 datasets. The questions in the study were resampled from a question pool of 50 per task so that each question would have the opportunity to be graded multiple times by multiple graders. The graders were asked to select, from each of the possible signals for that task, which one was conveyed in the sample. We presented our study to students at Arizona State University. All participants were fluent English speakers. 70\% were in STEM programs, 8\% were in humanities programs, and 22\% were in business or finance programs. 62\% were undergraduates, 16\% were in Master's programs, and 22\% were in PhD programs. We did not collect any other demographic information on participants. We collected 814 graded questions, 74 grades for each task. Among questions that had responses from three or more graders, our Fleiss's Kappa score was 0.85, indicating very strong agreement.

\subsection{Task Selection}

To identify tasks where expressivity could actually be perceived, we computed the normalized mutual information scores $N=I(\hat{s}; s)/H(s)$ between the grader's guesses and the ground truth signals. We rejected tasks where $N$ was below 0.1. This eliminated two tasks: profession and hobby. Our complete results for this task can be seen in Table \ref{tab:task-selection}. Of the tasks that ultimately made it through this selection process, none had an $N$ score below 0.15. 

\begin{table}[htb]\centering
\caption{Entropy scores for human graders across all tasks}\label{tab:task-selection}
\begin{tabular}{lrrrr}\toprule
Task &$I(s; \hat{s})$ (b) &$H(s)$ (b) &$N$ \\\midrule
tone &1.92 &1.99 &0.97 \\
skill level &0.72 &1.00 &0.72 \\
genre &0.78 &2.32 &0.34 \\
age &0.43 &1.42 &0.30 \\
register &0.26 &1.00 &0.26 \\
emotion &0.66 &2.78 &0.24 \\
MPAA rating &0.23 &1.21 &0.19 \\
gender &0.15 &0.78 &0.19 \\
political slant &0.22 &1.49 &0.15 \\
hobby &0.05 &1.63 &0.03 \\
profession &0.04 &1.44 &0.03 \\
\bottomrule
\end{tabular}
\end{table}

\subsection{Grader Selection}

We also used the results of our human study to identify which model would best serve as an automated grader. Our goal was to find a model which would behave in the most ``human-like'' manner; the one that would give the same response as a human most often. We selected graders this way rather than selecting the one that inferred the correct signal most often in order to capture a more human-centric picture of expressivity; a grader that is \emph{more} sensitive than a human at identifying expressive signals would give more forgiving scores to less-expressive models. This selection method also negates the possibility that models would be sensitive to signals in qualitatively different ways.

In order to measure each model's ``accuracy'' compared to human graders, we set ground truth labels for each of the unique texts that received a response in our human study. For texts that received more than one response, we accepted only the most-frequently answered one as correct. In the case of a tie, we accepted any of the tied signals as correct. We then applied 5 models in a grader setup to return their inferred signals. The 5 models were: GPT-4.1-2025-04-14, GPT-4o-2024-11-20, LLaMA 3.1-8b, Mistral v0.3-7b and Gemma 3-12b, hereafter referred to simply as GPT-4.1, GPT-4o, LLaMA 3.1, Mistral and Gemma 3. Models' responses were assigned a similarity score: the proportion of responses that matched the ``correct'' human label(s). Full similarity scores are reported in Table \ref{tab:model-selection}. Given that GPT-4o achieved the highest performance, we selected it as our grader.

\begin{table}[h]\centering
\small
\caption{Similarities of models tested to human answers}\label{tab:model-selection}
\begin{tabular}{p{4cm}rr}\toprule
Model &Similarity \\\midrule
GPT-4.1 &0.87 \\
GPT-4o &0.92 \\
LLaMA 3.1 &0.72 \\
Mistral &0.79\\
Gemma 3 &0.85 \\
\bottomrule
\end{tabular}
\end{table}

\section{Evaluation Results}

\begin{table}[!htp]\centering
\caption{Baseline $N$ scores for human-written text used for normalization}\label{tab:human-baseline}
\small
\begin{tabular}{p{4cm}rr}\toprule
Task &$N$ Score \\\midrule
MPAA rating &0.15 \\
age &0.30 \\
emotions &0.23 \\
gender &0.18 \\
genre &0.29 \\
political\_slant &0.15 \\
register &0.24 \\
skill &0.62 \\
tones &0.94 \\
\bottomrule
\end{tabular}
\end{table}

\begin{table*}[!htp]
\centering
\caption{Expressivity scores for each model tested across all tasks. Scores which indicate performance at or above human level have been set in \textbf{bold}. See Table \ref{tab:bootstrap-intervals} for bootstrap sampling.}
\label{tab:normscores}
\resizebox{\textwidth}{!}{%
\begin{tabular}{lrrrrrrrrrr}
\toprule
\multirow{2}{*}{Model} & \multicolumn{9}{c}{Expressivity Score} \\
\cmidrule{2-10}
& MPAA rating & age & emotions & gender & genre & political slant & register & skill & tones \\
\midrule
Gemma 2 & 0.88 & 0.73 & \textbf{2.94} & 0.08 & \textbf{1.83} & 0.14 & 0.00 & 0.38 & \textbf{1.01} \\
Gemma 3 & 0.92 & 0.96 & \textbf{2.91} & 0.14 & \textbf{1.89} & 0.14 & 0.78 & 0.36 & \textbf{1.04} \\
LLaMA 2 & \textbf{1.48} & 0.83 & \textbf{2.18} & 0.37 & \textbf{2.29} & 0.49 & 0.83 & 0.03 & 0.84 \\
LLaMA 3.1 & \textbf{2.28} & 0.57 & \textbf{2.50} & 0.54 & \textbf{2.43} & 0.31 & \textbf{1.05} & 0.16 & 0.99 \\
GPT-4.1 & \textbf{2.47} & 0.64 & \textbf{2.55} & 0.56 & \textbf{2.27} & 0.56 & \textbf{1.00} & 0.64 & \textbf{1.00} \\
GPT-4o & \textbf{2.29} & 0.65 & \textbf{3.32} & \textbf{1.34} & \textbf{3.22} & 0.92 & 0.82 & 0.05 & \textbf{1.06} \\
Mistral & \textbf{3.67} & \textbf{1.06} & \textbf{3.00} & \textbf{1.21} & \textbf{2.87} & 0.38 & 0.47 & 0.00 & 0.97 \\
\bottomrule
\end{tabular}}
\end{table*}

\begin{figure*}
    \centering
    \caption{Raw mutual information scores $I(\hat{s}; s)$ for each model across different ExpressivityBench tasks.}

    \includegraphics[width=0.93\linewidth]{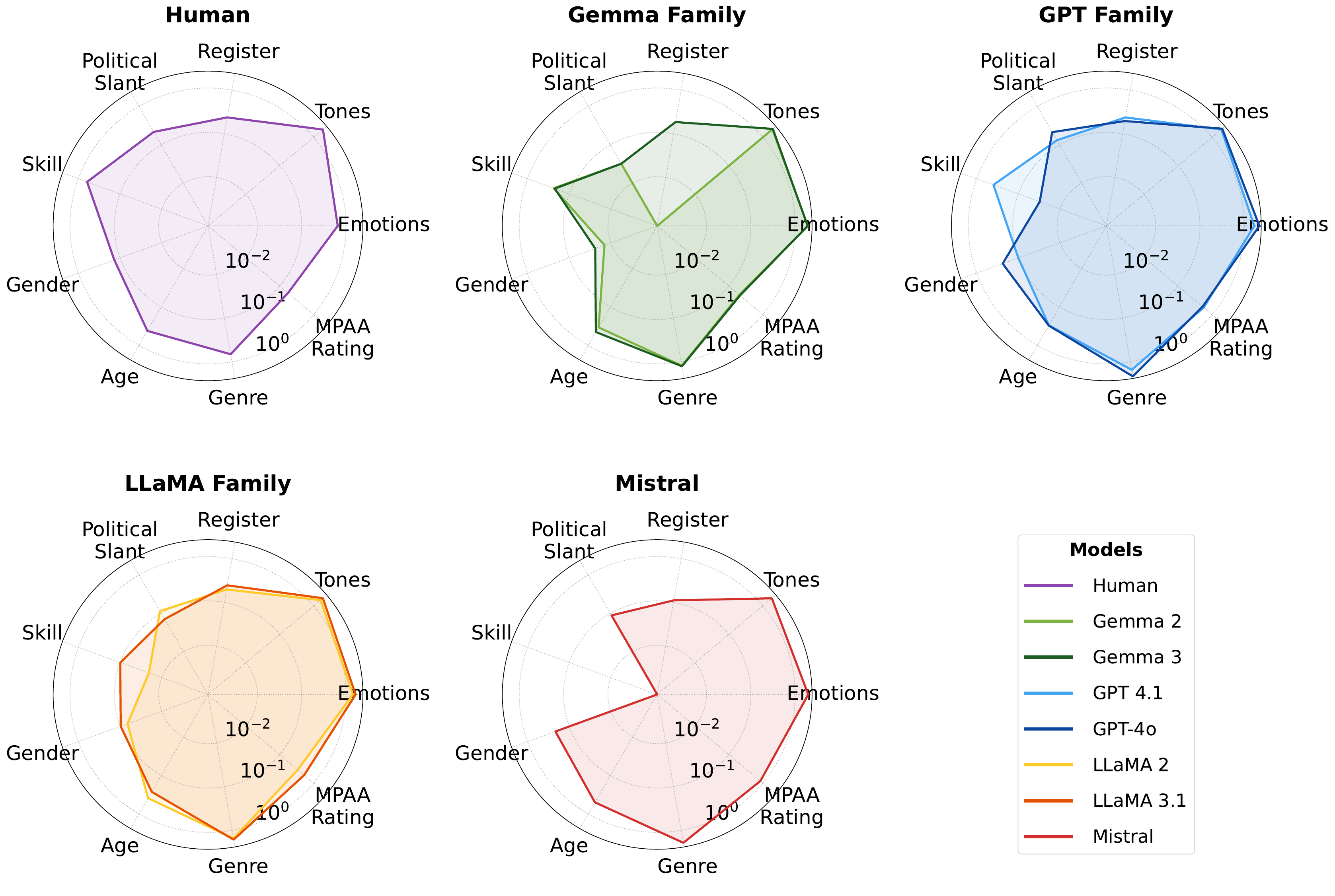}
    \label{fig:entropy-scores}
\end{figure*}

We evaluated 7 models using ExpressivityBench. These were LLaMA 3.1, GPT-4.1, GPT-4o, Mistral, and Gemma 3, as well as LLaMA 2-7b and Gemma 2-9b, hereafter referred to as LLaMA 2 and Gemma 2. We computed two scores per task per model: one raw mutual information score, and one human-normalized score. Each model was used to generate 50 samples per label per task for evaluation; this was 1550 samples in all. To compute a score normalized to human performance, we passed our human-written text through our benchmark to compute mutual entropy scores. Note that these differ from the scores computed in Table \ref{tab:task-selection}, as those scores were given by human graders. For consistency, we computed human baseline scores using the GPT-4o automated grader. The normalized $N$ scores for human-written text on each task can be seen in Table \ref{tab:human-baseline}. Models' mutual information scores were then divided by these human scores to get a baseline normalized for human performance. We dubbed these final values ``expressivity scores.'' A value of 1 indicates a performance on par with human expressivity in a given task; higher values indicate more expressivity. The complete results, in expressivity score, for each model can be seen in Table \ref{tab:normscores}. The unnormalized mutual information scores can be seen in Figure \ref{fig:entropy-scores}.

\section{Discussion}

Our evaluation using ExpressivityBench reveals that modern language models have mixed capabilities in generating stylistically expressive language, completely failing some tasks while outperforming human-written text in others. Specifically, tasks like MPAA rating and genre identification saw multiple models achieve scores exceeding human baselines by large margins. However, this trend does not generalize: in tasks rooted in identity expression, such as political slant, age, gender, and skill, most models performed significantly below human levels. No model consistently excelled across both stylistic and sociolinguistic dimensions.

Interestingly, two of the easiest tasks for models, MPAA ratings and genre, were both related to narrative structures. Tasks centered on stylistic genre mimicry leverage overt linguistic cues and benefit from extensive training data. These tasks often involve clear markers like specific vocabulary and sentence structures, enabling models to excel by amplifying these features [Example \ref{ex:genre}] \cite{cuevas2021metadiscursive}. Tone and emotion are often communicated through similar syntax- or lexicon-level stylistic markers, perhaps explaining LLM overperformance in these domains as well \cite{orekhova2022linguistic}. For instance, instruction-tuned models have been shown to overuse certain stylistic elements, leading to outputs that are more formulaic and less nuanced than human-authored texts \cite{Reinhart_2025}.

In contrast, tasks that require understanding and conveying implied identity (e.g. political slant, skill level, gender, age) demand a deeper grasp of social context, cultural nuances, and interpersonal dynamics \cite{andersen2001pragmatic}. Current LLMs often struggle with these tasks, as they lack the social awareness and contextual grounding that humans naturally possess \cite{yang2025sociallyawarelanguagetechnologies}. These results show that most LLMs are capable of surface-level expressivity, conveying emotions or tones quite well, but will falter when given more persona-oriented generation tasks, where an LLM must more fully emulate a human [Example \ref{ex:politics}]. 

Among the models tested, Mistral and GPT-4o stand out for their expressivity, especially in narrative and emotional dimensions. They are the only two models to score above human performance in five different tasks. However, their performance collapses on more subtle pragmatic tasks such as register and skill, where Mistral's score in particular falls to 0.00. Mistral and GPT-4o are among the worst in these tasks, suggesting that high performance in very broad kinds of signals, such as emotion and tone, may come at a cost of expressivity in these more nuanced tasks. Indeed, the three highest-performing models on the register task (LLaMA 2, LLaMA 3.1, and GPT-4.1) are also the three lowest-performing models on the emotions task. Gemma 2 and 3 are among the least expressive models across most dimensions, with particularly low scores in gender and political slant. 

One surprising outcome is that on some tasks, models may be vastly more expressive than humans. For example, in the emotion task, GPT-4o and Mistral achieved scores over triple the human baseline. Yet at that level, it is unclear whether overuse of emotional keywords may indeed make a text \emph{less} humanesque and natural [Example \ref{ex:emotions}]. Particularly given that models which perform excessively high on the emotions task seem to have poorer performance on others, like register and skill level, it may be that extremely high performance on any given task is undesirable. Future research could integrate human evaluation of LLM-generated text to explore whether there is a region of hyper-expressivity above which text no longer feels natural to readers.

\section{Conclusion}

This study introduces ExpressivityBench, a comprehensive benchmark designed to evaluate language models' ability to generate expressive, human-like text across multiple dimensions. Our evaluation spanned nine diverse tasks, assessing models on attributes ranging from emotion and tone to age, gender, and political slant. We tested a range of state-of-the-art models, including GPT-4.1, Mistral, LLaMA 3.1. Each task was paired with a human-authored control, allowing for direct comparisons between model output and natural language usage. Results show that while LLMs are capable of high performance in certain expressive tasks, their strengths are uneven and deeply task-dependent.

Models generally excelled in tasks related to emotion and narrative style, where surface-level cues and abundant training data provide strong signals. In these domains, models such as GPT-4o and Mistral even surpassed human-written baselines. However, the same models performed significantly worse on tasks requiring a nuanced grasp of implied identity, such as political slant, skill level, and register. These findings support a key insight: expressivity alone is not sufficient for human-like communication. While models can mimic overt stylistic traits, they struggle to balance expressivity with contextual appropriateness, which is essential for conveying identity and social meaning. Furthermore, our findings raise the possibility of ``hyper-expressivity,'' where excessive signaling of attributes like emotion leads to outputs that read as unnatural or inauthentic. As models become more expressive, future research must extend this work to understand how human readers react to this level of expressivity.

\section{Limitations}

While our study offers a new framework for evaluating implicit communication in LLMs, we recognize several limitations and provide responses below.

Using LLMs to evaluate other LLMs raises valid concerns about evaluation bias and circularity. To address this, we validate our graders against human annotations collected through a controlled user study. We also compute inter-rater reliability between human and model-based evaluators, finding strong agreement. We additionally follow \citet{laskar-etal-2024-systematic}'s best practices with regards to sharing grader versioning details and underlying data, which are provided alongside our paper. Human baselines further contextualize performance to avoid overreliance on automated graders.

We agree that implicit meaning varies across individuals and contexts. However, our approach operationalizes expressivity using an information-theoretic framework: we define expressivity as the reduction in uncertainty about an intended target property given a model’s output. This enables us to quantify expressivity as a measurable communicative effect, not just a stylistic artifact. Our methodology is inspired by communication theory and recent work in computational pragmatics, and supports consistent evaluation across a range of tasks while allowing for human-grounded variation.

This initial version of ExpressivityBench focuses on English-language expressivity, which may not reflect norms or phenomena in other cultures and languages. However, the benchmark is designed to be modular and extensible. Many of our tasks, including tone modulation and identity-marked variation, have analogues in other linguistic and cultural settings. Future work could explore extending ExpressivityBench with multilingual LLMs in collaboration with domain experts to ensure cultural validity

\section{Ethical Statement}

Since our paper is a generic algorithmic evaluation, we do not foresee direct negative societal impacts. Our study was ruled exempt by the Arizona State University IRB. Human graders who were surveyed for experiment 1 were all given a privacy statement notifying them of their confidentiality and of the purpose of the experiment. No identifying information was solicited or collected. Information given to survey participants on confidentiality can be found in Appendix \ref{human-study}.

\bibliography{bibliography}

@article{bai2024benchmarking,
  title={Benchmarking foundation models with language-model-as-an-examiner},
  author={Bai, Yushi and Ying, Jiahao and Cao, Yixin and Lv, Xin and He, Yuze and Wang, Xiaozhi and Yu, Jifan and Zeng, Kaisheng and Xiao, Yijia and Lyu, Haozhe and others},
  journal={Advances in Neural Information Processing Systems},
  volume={36},
  year={2024}
}

@article{chern2024can,
  title={Can Large Language Models be Trusted for Evaluation? Scalable Meta-Evaluation of LLMs as Evaluators via Agent Debate},
  author={Chern, Steffi and Chern, Ethan and Neubig, Graham and Liu, Pengfei},
  journal={arXiv preprint arXiv:2401.16788},
  year={2024},
}

@article{benedetto2013readers,
  title={E-readers and visual fatigue},
  author={Benedetto, Simone and Drai-Zerbib, V{\'e}ronique and Pedrotti, Marco and Tissier, Geoffrey and Baccino, Thierry},
  journal={PloS one},
  volume={8},
  number={12},
  pages={e83676},
  year={2013},
  publisher={Public Library of Science San Francisco, USA}
}

@article{lee2023rlaif,
  title={Rlaif: Scaling reinforcement learning from human feedback with ai feedback},
  author={Lee, Harrison and Phatale, Samrat and Mansoor, Hassan and Lu, Kellie and Mesnard, Thomas and Bishop, Colton and Carbune, Victor and Rastogi, Abhinav},
  journal={arXiv preprint arXiv:2309.00267},
  year={2023}
}

@article{OpenAI2023GPT4TR,
  author = {OpenAI},
  description = {This paper reports the development of GPT-4, a large-scale, multimodal model which can accept image and text inputs and produce text outputs. GPT-4 exhibits human-level performance on various professional and academic benchmarks, including passing a simulated bar exam with a score around the top 10% of test takers.},
  journal = {ArXiv},
  title = {GPT-4 Technical Report},
  volume = {abs/2303.08774},
  year = 2023
}

@article{brown2020language,
  author = {Brown, Tom B and Mann, Benjamin and Ryder, Nick and Subbiah, Melanie and Kaplan, Jared and Dhariwal, Prafulla and Neelakantan, Arvind and Shyam, Pranav and Sastry, Girish and Askell, Amanda and others},
  journal = {arXiv preprint arXiv:2005.14165},
  title = {Language models are few-shot learners},
  year = 2020
}

@misc{radford_language_2019,
  author = {Radford, Alec and Wu, Jeff and Child, Rewon and Luan, D. and Amodei, Dario and Sutskever, Ilya},
  title = {Language models are unsupervised multitask learners},
  urldate = {2023-01-06},
  year = 2019
}

@misc{lin2023llmeval,
      title={LLM-Eval: Unified Multi-Dimensional Automatic Evaluation for Open-Domain Conversations with Large Language Models}, 
      author={Yen-Ting Lin and Yun-Nung Chen},
      year={2023},
      archivePrefix={arXiv},
      primaryClass={cs.CL}
}

@misc{touvron2023llama,
  author = {Touvron, Hugo and Lavril, Thibaut and Izacard, Gautier and Martinet, Xavier and Lachaux, Marie-Anne and Lacroix, Timothée and Rozière, Baptiste and Goyal, Naman and Hambro, Eric and Azhar, Faisal and Rodriguez, Aurelien and Joulin, Armand and Grave, Edouard and Lample, Guillaume},
  description = {[2302.13971] LLaMA: Open and Efficient Foundation Language Models},
  note = {cite arxiv:2302.13971},
  title = {LLaMA: Open and Efficient Foundation Language Models},
  year = 2023
}

@inproceedings{plaza-emotion-analysis,
    title = "Emotion Analysis in {NLP}: Trends, Gaps and Roadmap for Future Directions",
    author = "Plaza-del-Arco, Flor Miriam  and
      Cercas Curry, Alba A.  and
      Cercas Curry, Amanda  and
      Hovy, Dirk",
    editor = "Calzolari, Nicoletta  and
      Kan, Min-Yen  and
      Hoste, Veronique  and
      Lenci, Alessandro  and
      Sakti, Sakriani  and
      Xue, Nianwen",
    booktitle = "Proceedings of the 2024 Joint International Conference on Computational Linguistics, Language Resources and Evaluation (LREC-COLING 2024)",
    month = may,
    year = "2024",
    address = "Torino, Italia",
    publisher = "ELRA and ICCL",
    pages = "5696--5710",
}

@article{sharma2024critical,
  title={A Critical Evaluation of AI Feedback for Aligning Large Language Models},
  author={Sharma, Archit and Keh, Sedrick and Mitchell, Eric and Finn, Chelsea and Arora, Kushal and Kollar, Thomas},
  journal={arXiv preprint arXiv:2402.12366},
  year={2024}
}

@article{tunstall2023zephyr,
  title={Zephyr: Direct distillation of lm alignment},
  author={Tunstall, Lewis and Beeching, Edward and Lambert, Nathan and Rajani, Nazneen and Rasul, Kashif and Belkada, Younes and Huang, Shengyi and von Werra, Leandro and Fourrier, Cl{\'e}mentine and Habib, Nathan and others},
  journal={arXiv preprint arXiv:2310.16944},
  year={2023}
}

@inbook{EssentialsOfLinguistics,
    author = {Catherine Anderson; Bronwyn Bjorkman; Derek Denis; Julianne Doner; Margaret Grant; Nathan Sanders; and Ai Taniguchi},
    title = {Essentials of Linguistics},
    publisher = {eCampusOntario},
    year = {2022},
    chapter = 7
}

@article{ModernLinguisticExpressiveness,
    author = {Margaret Apresyan},
    title = {On The Concept of “Expressiveness” In Modern Linguistics},
    journal = {Annals of Language and Literature},
    year = 2018
}

@article{devlin2018bert,
  title={Bert: Pre-training of deep bidirectional transformers for language understanding},
  author={Devlin, Jacob and Chang, Ming-Wei and Lee, Kenton and Toutanova, Kristina},
  journal={arXiv preprint arXiv:1810.04805},
  year={2018}
}

@article{collins2023evaluating,
  title={Evaluating language models for mathematics through interactions},
  author={Collins, Katherine M and Jiang, Albert Q and Frieder, Simon and Wong, Lionel and Zilka, Miri and Bhatt, Umang and Lukasiewicz, Thomas and Wu, Yuhuai and Tenenbaum, Joshua B and Hart, William and others},
  journal={arXiv preprint arXiv:2306.01694},
  year={2023}
}

@misc{verga2024replacing,
      title={Replacing Judges with Juries: Evaluating LLM Generations with a Panel of Diverse Models}, 
      author={Pat Verga and Sebastian Hofstatter and Sophia Althammer and Yixuan Su and Aleksandra Piktus and Arkady Arkhangorodsky and Minjie Xu and Naomi White and Patrick Lewis},
      year={2024},
      eprint={2404.18796},
      archivePrefix={arXiv},
      primaryClass={cs.CL},
      url={https://arxiv.org/abs/2404.18796}, 
}

@misc{chiang2024chatbot,
      title={Chatbot Arena: An Open Platform for Evaluating LLMs by Human Preference}, 
      author={Wei-Lin Chiang and Lianmin Zheng and Ying Sheng and Anastasios Nikolas Angelopoulos and Tianle Li and Dacheng Li and Hao Zhang and Banghua Zhu and Michael Jordan and Joseph E. Gonzalez and Ion Stoica},
      year={2024},
      archivePrefix={arXiv},
      primaryClass={cs.AI}
}

@article{surveyevalllm,
author = {Chang, Yupeng and Wang, Xu and Wang, Jindong and Wu, Yuan and Yang, Linyi and Zhu, Kaijie and Chen, Hao and Yi, Xiaoyuan and Wang, Cunxiang and Wang, Yidong and Ye, Wei and Zhang, Yue and Chang, Yi and Yu, Philip S. and Yang, Qiang and Xie, Xing},
title = {A Survey on Evaluation of Large Language Models},
year = {2024},
issue_date = {June 2024},
publisher = {Association for Computing Machinery},
address = {New York, NY, USA},
volume = {15},
number = {3},
issn = {2157-6904},
journal = {ACM Trans. Intell. Syst. Technol.},
month = {mar},
articleno = {39},
numpages = {45}
}

@inproceedings{dai2023can,
  title={Can large language models provide feedback to students? A case study on ChatGPT},
  author={Dai, Wei and Lin, Jionghao and Jin, Hua and Li, Tongguang and Tsai, Yi-Shan and Ga{\v{s}}evi{\'c}, Dragan and Chen, Guanliang},
  booktitle={2023 IEEE International Conference on Advanced Learning Technologies (ICALT)},
  pages={323--325},
  year={2023},
  organization={IEEE}
}

@misc{bubeck2023sparks,
      title={Sparks of Artificial General Intelligence: Early experiments with GPT-4}, 
      author={Sébastien Bubeck and Varun Chandrasekaran and Ronen Eldan and Johannes Gehrke and Eric Horvitz and Ece Kamar and Peter Lee and Yin Tat Lee and Yuanzhi Li and Scott Lundberg and Harsha Nori and Hamid Palangi and Marco Tulio Ribeiro and Yi Zhang},
      year={2023},
      archivePrefix={arXiv},
      primaryClass={cs.CL}
}

@misc{parmar2024logicbench,
      title={LogicBench: Towards Systematic Evaluation of Logical Reasoning Ability of Large Language Models}, 
      author={Mihir Parmar and Nisarg Patel and Neeraj Varshney and Mutsumi Nakamura and Man Luo and Santosh Mashetty and Arindam Mitra and Chitta Baral},
      year={2024},
      archivePrefix={arXiv},
      primaryClass={cs.CL}
}

@article{demszky2020goemotions,
  title={GoEmotions: A dataset of fine-grained emotions},
  author={Demszky, Dorottya and Movshovitz-Attias, Dana and Ko, Jeongwoo and Cowen, Alan and Nemade, Gaurav and Ravi, Sujith},
  journal={arXiv preprint arXiv:2005.00547},
  year={2020},
}

@article{yus1999misunderstandings,
  title={Misunderstandings and explicit/implicit communication},
  author={Yus, Francisco},
  journal={Pragmatics. Quarterly Publication of the International Pragmatics Association (IPrA)},
  volume={9},
  number={4},
  pages={487--517},
  year={1999},
  publisher={John Benjamins Publishing Company Amsterdam/Philadephia}
}

@inproceedings{zhu-etal-2024-large,
    title = "Can Large Language Models Understand Context?",
    author = "Zhu, Yilun  and
      Moniz, Joel Ruben Antony  and
      Bhargava, Shruti  and
      Lu, Jiarui  and
      Piraviperumal, Dhivya  and
      Li, Site  and
      Zhang, Yuan  and
      Yu, Hong  and
      Tseng, Bo-Hsiang",
    editor = "Graham, Yvette  and
      Purver, Matthew",
    booktitle = "Findings of the Association for Computational Linguistics: EACL 2024",
    month = mar,
    year = "2024",
    address = "St. Julian{'}s, Malta",
    publisher = "Association for Computational Linguistics",
    url = "https://aclanthology.org/2024.findings-eacl.135",
    pages = "2004--2018",
}

@inproceedings{knepper2017implicit,
author = {Knepper, Ross A. and Mavrogiannis, Christoforos I. and Proft, Julia and Liang, Claire},
title = {Implicit Communication in a Joint Action},
year = {2017},
isbn = {9781450343367},
publisher = {Association for Computing Machinery},
address = {New York, NY, USA},
url = {https://doi.org/10.1145/2909824.3020226},
doi = {10.1145/2909824.3020226},
booktitle = {Proceedings of the 2017 ACM/IEEE International Conference on Human-Robot Interaction},
pages = {283–292},
numpages = {10},
location = {Vienna, Austria},
series = {HRI '17}
}

@inproceedings{trustllm,
  title={TrustLLM: Trustworthiness in Large Language Models},
  author={Huang, Yue and Sun, Lichao and Wang, Haoran and Wu, Siyuan and Zhang, Qihui and Li, Yuan and Gao, Chujie and Huang, Yixin and Lyu, Wenhan and Zhang, Yixuan and others},
  booktitle={International Conference on Machine Learning},
  pages={20166--20270},
  year={2024},
  organization={PMLR}
}

@book{green2016slang,
  title={Slang: A very short introduction},
  author={Green, Jonathon},
  volume={465},
  year={2016},
  publisher={Oxford University Press}
}

@inproceedings{gu-etal-2022-just,
    title = "Just-{DREAM}-about-it: Figurative Language Understanding with {DREAM}-{FLUTE}",
    author = "Gu, Yuling  and
      Fu, Yao  and
      Pyatkin, Valentina  and
      Magnusson, Ian  and
      Dalvi Mishra, Bhavana  and
      Clark, Peter",
    editor = "Ghosh, Debanjan  and
      Beigman Klebanov, Beata  and
      Muresan, Smaranda  and
      Feldman, Anna  and
      Poria, Soujanya  and
      Chakrabarty, Tuhin",
    booktitle = "Proceedings of the 3rd Workshop on Figurative Language Processing (FLP)",
    month = dec,
    year = "2022",
    address = "Abu Dhabi, United Arab Emirates (Hybrid)",
    publisher = "Association for Computational Linguistics",
    url = "https://aclanthology.org/2022.flp-1.12",
    doi = "10.18653/v1/2022.flp-1.12",
    pages = "84--93",
}

@inproceedings{chakrabarty-etal-2022-flute,
    title = "{FLUTE}: Figurative Language Understanding through Textual Explanations",
    author = "Chakrabarty, Tuhin  and
      Saakyan, Arkadiy  and
      Ghosh, Debanjan  and
      Muresan, Smaranda",
    editor = "Goldberg, Yoav  and
      Kozareva, Zornitsa  and
      Zhang, Yue",
    booktitle = "Proceedings of the 2022 Conference on Empirical Methods in Natural Language Processing",
    month = dec,
    year = "2022",
    address = "Abu Dhabi, United Arab Emirates",
    publisher = "Association for Computational Linguistics",
    url = "https://aclanthology.org/2022.emnlp-main.481",
    doi = "10.18653/v1/2022.emnlp-main.481",
    pages = "7139--7159",
}

@inproceedings{li-etal-2017-dailydialog,
    title = "{D}aily{D}ialog: A Manually Labelled Multi-turn Dialogue Dataset",
    author = "Li, Yanran  and
      Su, Hui  and
      Shen, Xiaoyu  and
      Li, Wenjie  and
      Cao, Ziqiang  and
      Niu, Shuzi",
    editor = "Kondrak, Greg  and
      Watanabe, Taro",
    booktitle = "Proceedings of the Eighth International Joint Conference on Natural Language Processing (Volume 1: Long Papers)",
    month = nov,
    year = "2017",
    address = "Taipei, Taiwan",
    publisher = "Asian Federation of Natural Language Processing",
    url = "https://aclanthology.org/I17-1099",
    pages = "986--995",
}

@inproceedings{choi-etal-2023-llms,
    title = "Do {LLM}s Understand Social Knowledge? Evaluating the Sociability of Large Language Models with {S}oc{KET} Benchmark",
    author = "Choi, Minje  and
      Pei, Jiaxin  and
      Kumar, Sagar  and
      Shu, Chang  and
      Jurgens, David",
    editor = "Bouamor, Houda  and
      Pino, Juan  and
      Bali, Kalika",
    booktitle = "Proceedings of the 2023 Conference on Empirical Methods in Natural Language Processing",
    month = dec,
    year = "2023",
    address = "Singapore",
    publisher = "Association for Computational Linguistics",
    url = "https://aclanthology.org/2023.emnlp-main.699",
    doi = "10.18653/v1/2023.emnlp-main.699",
    pages = "11370--11403",
}

@inproceedings{rashkin-etal-2019-towards,
    title = "Towards Empathetic Open-domain Conversation Models: A New Benchmark and Dataset",
    author = "Rashkin, Hannah  and
      Smith, Eric Michael  and
      Li, Margaret  and
      Boureau, Y-Lan",
    editor = "Korhonen, Anna  and
      Traum, David  and
      M{\`a}rquez, Llu{\'\i}s",
    booktitle = "Proceedings of the 57th Annual Meeting of the Association for Computational Linguistics",
    month = jul,
    year = "2019",
    address = "Florence, Italy",
    publisher = "Association for Computational Linguistics",
    url = "https://aclanthology.org/P19-1534",
    doi = "10.18653/v1/P19-1534",
    pages = "5370--5381",
}

@inproceedings{hu-etal-2023-fine,
    title = "A fine-grained comparison of pragmatic language understanding in humans and language models",
    author = "Hu, Jennifer  and
      Floyd, Sammy  and
      Jouravlev, Olessia  and
      Fedorenko, Evelina  and
      Gibson, Edward",
    editor = "Rogers, Anna  and
      Boyd-Graber, Jordan  and
      Okazaki, Naoaki",
    booktitle = "Proceedings of the 61st Annual Meeting of the Association for Computational Linguistics (Volume 1: Long Papers)",
    month = jul,
    year = "2023",
    address = "Toronto, Canada",
    publisher = "Association for Computational Linguistics",
    url = "https://aclanthology.org/2023.acl-long.230",
    doi = "10.18653/v1/2023.acl-long.230",
    pages = "4194--4213",
}

@article{ding2025citations,
  title={Citations and Trust in LLM Generated Responses},
  author={Ding, Yifan and Facciani, Matthew and Poudel, Amrit and Joyce, Ellen and Aguinaga, Salvador and Veeramani, Balaji and Bhattacharya, Sanmitra and Weninger, Tim},
  journal={arXiv preprint arXiv:2501.01303},
  year={2025}
}

@article{davies2023competence,
  title={Competence-based analysis of language models},
  author={Davies, Adam and Jiang, Jize and Zhai, ChengXiang},
  journal={arXiv preprint arXiv:2303.00333},
  year={2023}
}

@inproceedings{ziyu2023through,
  title={Through the lens of core competency: Survey on evaluation of large language models},
  author={Ziyu, Zhuang and Qiguang, Chen and Longxuan, Ma and Mingda, Li and Yi, Han and Yushan, Qian and Haopeng, Bai and Weinan, Zhang and Liu, Ting},
  booktitle={Proceedings of the 22nd Chinese National Conference on Computational Linguistics (Volume 2: Frontier Forum)},
  pages={88--109},
  year={2023}
}

@article{hao2024llm,
  title={LLM Reasoners: New Evaluation, Library, and Analysis of Step-by-Step Reasoning with Large Language Models},
  author={Hao, Shibo and Gu, Yi and Luo, Haotian and Liu, Tianyang and Shao, Xiyan and Wang, Xinyuan and Xie, Shuhua and Ma, Haodi and Samavedhi, Adithya and Gao, Qiyue and others},
  journal={arXiv preprint arXiv:2404.05221},
  year={2024}
}

@article{ferguson2025chatgpt,
  author = {Mackenzie Ferguson},
  title = {OpenAI Unleashes ChatGPT's Personality Customization: Now 'Gen Z', 'Chatty', or Whatever You Want!},
  journal = {OpenTools},
  year = {2025},
  month = {January},
  url = {https://opentools.com/article/chatgpt-personality-customization},
  note = {Last updated: 1/10/2025}
}

@article{hu2024recent,
  title={Recent trends of multimodal affective computing: A survey from NLP perspective},
  author={Hu, Guimin and Xin, Yi and Lyu, Weimin and Huang, Haojian and Sun, Chang and Zhu, Zhihong and Gui, Lin and Cai, Ruichu and Cambria, Erik and Seifi, Hasti},
  journal={arXiv preprint arXiv:2409.07388},
  year={2024}
}

@article{pyreddy2025emoxpt,
  title={EmoXpt: Analyzing Emotional Variances in Human Comments and LLM-Generated Responses},
  author={Pyreddy, Shireesh Reddy and Zaman, Tarannum Shaila},
  journal={arXiv preprint arXiv:2501.06597},
  year={2025}
}

@misc{shi2023redteaminglanguagemodel,
      title={Red Teaming Language Model Detectors with Language Models}, 
      author={Zhouxing Shi and Yihan Wang and Fan Yin and Xiangning Chen and Kai-Wei Chang and Cho-Jui Hsieh},
      year={2023},
      eprint={2305.19713},
      archivePrefix={arXiv},
      primaryClass={cs.CL},
      url={https://arxiv.org/abs/2305.19713}, 
}

@article{stecker2001expressiveness,
  title={Expressiveness and expression in music and poetry},
  author={Stecker, Robert},
  journal={The Journal of aesthetics and art criticism},
  volume={59},
  number={1},
  pages={85--96},
  year={2001},
  publisher={JSTOR}
}

@inproceedings{sreeja2019perc,
  title={Perc-an emotion recognition corpus for cognitive poems},
  author={Sreeja, PS and Mahalakshmi, GS},
  booktitle={2019 International Conference on Communication and Signal Processing (ICCSP)},
  pages={0200--0207},
  year={2019},
  organization={IEEE}
}

@article{pavlick2016empirical,
  title={An empirical analysis of formality in online communication},
  author={Pavlick, Ellie and Tetreault, Joel},
  journal={Transactions of the association for computational linguistics},
  volume={4},
  pages={61--74},
  year={2016},
  publisher={MIT Press One Rogers Street, Cambridge, MA 02142-1209, USA journals-info~…}
}

@inproceedings{crossley2021commonlit,
  title={The CommonLit Ease of Readability (CLEAR) Corpus.},
  author={Crossley, Scott A and Heintz, Aron and Choi, Joon Suh and Batchelor, Jordan and Karimi, Mehrnoush and Malatinszky, Agnes},
  booktitle={EDM},
  year={2021}
}

@misc{atif2023tone,
    author={Atif, Sameed},
    title={English sentence tone analysis},
    year={2023},
  howpublished = {\url{https://www.kaggle.com/datasets/sameedatif/tone-analysis}},

}

@article{crossley5129353large,
  title={A Large-Scale Corpus for Assessing Source-Based Writing Quality: Asap 2.0},
  author={Crossley, Scott Andrew and Baffour, Perpetual and Burleigh, L and King, Jules},
  journal={Available at SSRN 5129353}
}

@inproceedings{haak2023qbias,
  title={Qbias-A Dataset on Media Bias in Search Queries and Query Suggestions},
  author={Haak, Fabian and Schaer, Philipp},
  booktitle={Proceedings of the 15th ACM Web Science Conference 2023},
  pages={239--244},
  year={2023}
}

@inproceedings{schler2006blogging,
  author       = {Schler, Jonathan and Koppel, Moshe and Argamon, Shlomo and Pennebaker, James},
  title        = {Effects of Age and Gender on Blogging},
  booktitle    = {Proceedings of the 2006 AAAI Spring Symposium on Computational Approaches for Analyzing Weblogs},
  year         = {2006},
  url          = {http://www.cs.biu.ac.il/~schlerj/schler_springsymp06.pdf}
}

@misc{zhang2018personalizingdialogueagentsi,
      title={Personalizing Dialogue Agents: I have a dog, do you have pets too?}, 
      author={Saizheng Zhang and Emily Dinan and Jack Urbanek and Arthur Szlam and Douwe Kiela and Jason Weston},
      year={2018},
      eprint={1801.07243},
      archivePrefix={arXiv},
      primaryClass={cs.AI},
      url={https://arxiv.org/abs/1801.07243}, 
}

@misc{yang2025sociallyawarelanguagetechnologies,
      title={The Call for Socially Aware Language Technologies}, 
      author={Diyi Yang and Dirk Hovy and David Jurgens and Barbara Plank},
      year={2025},
      eprint={2405.02411},
      archivePrefix={arXiv},
      primaryClass={cs.CL},
      url={https://arxiv.org/abs/2405.02411}, 
}

@article{Reinhart_2025,
   title={Do LLMs write like humans? Variation in grammatical and rhetorical styles},
   volume={122},
   ISSN={1091-6490},
   url={http://dx.doi.org/10.1073/pnas.2422455122},
   DOI={10.1073/pnas.2422455122},
   number={8},
   journal={Proceedings of the National Academy of Sciences},
   publisher={Proceedings of the National Academy of Sciences},
   author={Reinhart, Alex and Markey, Ben and Laudenbach, Michael and Pantusen, Kachatad and Yurko, Ronald and Weinberg, Gordon and Brown, David West},
   year={2025},
   month=feb }

@article{orekhova2022linguistic,
  title={Linguistic and Stylistic Markers of Influence in the Essayistic Text: A Linguophilosophic Aspect},
  author={Orekhova, Larysa Ivanivna and Gremaliuk, Tetiana and Borysenko, Natalia and Cheban, Oksana},
  year={2022},
  publisher={International Journal of Computer Science and Network Security}
}

@article{andersen2001pragmatic,
  title={Pragmatic markers and sociolinguistic variation},
  author={Andersen, Gisle},
  year={2001},
  publisher={John Benjamins Publishing Company}
}

@article{cuevas2021metadiscursive,
  title={Metadiscursive markers and text genre: A metareview},
  author={Cuevas-Alonso, Miguel and M{\'\i}guez-{\'A}lvarez, Carla},
  journal={Publications},
  volume={9},
  number={4},
  pages={56},
  year={2021},
  publisher={MDPI}
}

@article{gutt1996implicit,
  title={Implicit information in literary translation: A relevance-theoretic perspective},
  author={Gutt, Ernst-August},
  journal={Target. International Journal of Translation Studies},
  volume={8},
  number={2},
  pages={239--256},
  year={1996},
  publisher={John Benjamins}
}

@article{deckert2023various,
  title={Various dimensions of expressivity},
  author={Deckert, Miko{\l}aj and Kosecki, Krzysztof},
  journal={Lodz Papers in Pragmatics},
  volume={19},
  number={1},
  pages={1--6},
  year={2023},
  publisher={De Gruyter}
}

@inproceedings{choi2020ten,
  title={Ten social dimensions of conversations and relationships},
  author={Choi, Minje and Aiello, Luca Maria and Varga, Kriszti{\'a}n Zsolt and Quercia, Daniele},
  booktitle={Proceedings of The Web Conference 2020},
  pages={1514--1525},
  year={2020}
}

@inproceedings{laskar-etal-2024-systematic,
    title = "A Systematic Survey and Critical Review on Evaluating Large Language Models: Challenges, Limitations, and Recommendations",
    author = "Laskar, Md Tahmid Rahman  and
      Alqahtani, Sawsan  and
      Bari, M Saiful  and
      Rahman, Mizanur  and
      Khan, Mohammad Abdullah Matin  and
      Khan, Haidar  and
      Jahan, Israt  and
      Bhuiyan, Amran  and
      Tan, Chee Wei  and
      Parvez, Md Rizwan  and
      Hoque, Enamul  and
      Joty, Shafiq  and
      Huang, Jimmy",
    editor = "Al-Onaizan, Yaser  and
      Bansal, Mohit  and
      Chen, Yun-Nung",
    booktitle = "Proceedings of the 2024 Conference on Empirical Methods in Natural Language Processing",
    month = nov,
    year = "2024",
    address = "Miami, Florida, USA",
    publisher = "Association for Computational Linguistics",
    url = "https://aclanthology.org/2024.emnlp-main.764/",
    doi = "10.18653/v1/2024.emnlp-main.764",
    pages = "13785--13816"
}

@misc{yao2023colliesystematicconstructionconstrained,
      title={COLLIE: Systematic Construction of Constrained Text Generation Tasks}, 
      author={Shunyu Yao and Howard Chen and Austin W. Hanjie and Runzhe Yang and Karthik Narasimhan},
      year={2023},
      eprint={2307.08689},
      archivePrefix={arXiv},
      primaryClass={cs.CL},
      url={https://arxiv.org/abs/2307.08689}, 
}

@misc{chen2024evaluatingunderstandingimprovingconstrained,
      title={Evaluating, Understanding, and Improving Constrained Text Generation for Large Language Models}, 
      author={Xiang Chen and Xiaojun Wan},
      year={2024},
      eprint={2310.16343},
      archivePrefix={arXiv},
      primaryClass={cs.CL},
      url={https://arxiv.org/abs/2310.16343}, 
}

@misc{garbacea2025constrainedneurallanguagegeneration,
      title={Why is constrained neural language generation particularly challenging?}, 
      author={Cristina Garbacea and Qiaozhu Mei},
      year={2025},
      eprint={2206.05395},
      archivePrefix={arXiv},
      primaryClass={cs.CL},
      url={https://arxiv.org/abs/2206.05395}, 
}

@inproceedings{liu2024we,
  title={" We Need Structured Output": Towards User-centered Constraints on Large Language Model Output},
  author={Liu, Michael Xieyang and Liu, Frederick and Fiannaca, Alexander J and Koo, Terry and Dixon, Lucas and Terry, Michael and Cai, Carrie J},
  booktitle={Extended Abstracts of the CHI Conference on Human Factors in Computing Systems},
  pages={1--9},
  year={2024}
}

@article{greene2023teaching,
  title={Teaching Tone, Mood and Purpose through the Interpretation of Acti ist Poetr},
  author={Greene, Wm Miles},
  year={2023}
}

@misc{ma2025pragmaticseralargelanguage,
      title={Pragmatics in the Era of Large Language Models: A Survey on Datasets, Evaluation, Opportunities and Challenges}, 
      author={Bolei Ma and Yuting Li and Wei Zhou and Ziwei Gong and Yang Janet Liu and Katja Jasinskaja and Annemarie Friedrich and Julia Hirschberg and Frauke Kreuter and Barbara Plank},
      year={2025},
      eprint={2502.12378},
      archivePrefix={arXiv},
      primaryClass={cs.CL},
      url={https://arxiv.org/abs/2502.12378}, 
}

@misc{sravanthi2024pubpragmaticsunderstandingbenchmark,
      title={PUB: A Pragmatics Understanding Benchmark for Assessing LLMs' Pragmatics Capabilities}, 
      author={Settaluri Lakshmi Sravanthi and Meet Doshi and Tankala Pavan Kalyan and Rudra Murthy and Pushpak Bhattacharyya and Raj Dabre},
      year={2024},
      eprint={2401.07078},
      archivePrefix={arXiv},
      primaryClass={cs.CL},
      url={https://arxiv.org/abs/2401.07078}, 
}

@article{li2025dr,
  title={Dr Genre: Reinforcement Learning from Decoupled LLM Feedback for Generic Text Rewriting},
  author={Li, Yufei and Nham, John and Jawahar, Ganesh and Shu, Lei and Uthus, David and Sung, Yun-Hsuan and Yang, Chengrun and Rolnick, Itai and Qiao, Yi and Liu, Cong},
  journal={arXiv preprint arXiv:2503.06781},
  year={2025}
}

\section{Appendix}

\subsection{Example Generations}

\begin{enumerate}
    \item \label{ex:genre}
    \textbf{Amplified stylistic markers in genre tasks.}  
    Over-stylization can yield high expressivity but low naturalness:
    \begin{quote}
        \textit{(GPT-4o, genre = science fiction)} ``When the twin suns rose over the drowned megacity, Mira’s reflection in the nanoglass tide wasn’t her own—it was a quantum echo of someone who hadn’t yet been instantiated in the chrono-net.''  
        \textit{(Human example, genre = science fiction)} ``Levity is a universal constant. Comedy is one of the basic forces of the Universe. Mankind latches onto comedy, because levity is the expanding principle that keeps the whole bubble inflating.''
    \end{quote}
    The model exaggerates genre cues such as imagery and lexical intensity.
    \item \label{ex:politics}

    \begin{quote}
        \textit{(Gemma 3, political slant = left, excerpt)} ``Government waste must stop, but compassion matters''
        
    \end{quote}
    This sample juxtaposed two viewpoints to add nuance, but was graded as being on the left rather than the right. This is an instance where the LLM produced text which misapplied common talking points to realistically create a persona. It's possible that the model qualified its point to seem more human, but ended up undermining its position.

    \item \label{ex:emotions}
    \begin{quote}
        \emph{(GPT-4o, emotion = joy)} ``My heart bursts with radiant sunshine and glittering laughter spilling through every pore.''
    \end{quote}
    This is an example where the model produces text which is repetitive in its use of emotion-signals. In particular, constructions like ``radiant sunshine'' are redundant, and ``laughter'' spilling out through pores is nonstandard--laughter is not usually said to come out of pores.
\end{enumerate}

\subsection{Goodreads Genre Data \label{goodreads}}
Goodreads is a widely used social cataloging platform where users can track their reading, review books, and assign genre-specific tags. These user-generated tags form a rich and large-scale dataset that offers insight into how readers perceive genre distinctions. For our study, we leveraged Goodreads' genre tagging system to construct a representative dataset of literary text associated with specific genres.

We began by examining Goodreads’ publicly listed genres and identifying the ten genres with the highest number of user-tagged books. To focus our analysis on the most semantically distinct categories, we then applied the winnowing algorithm described in Appendix \ref{winnowing} to this top-ten list. This process yielded five genres with the most differentiated language patterns: ``science fiction,'' ``thriller,'' ``romance,'' ``humor,'' and ``fantasy.''

For each of these five genres, we used Goodreads’ internal search functionality to retrieve a list of books tagged exclusively with the target genre, excluding any works that also bore tags from the remaining four target genres. From within the top 1000 search results for each genre, we selected books at random. For each selected book, we attempted to extract a quote of between 3 and 7 sentences from the user-contributed ``Quotes'' section. Quotes were eliminated if they contained a genre name. If no such quote was available, the book was discarded, and another was sampled. This process was repeated iteratively until we had collected 60 qualifying quotes for each genre, yielding a total corpus of 300 genre-specific excerpts.

\subsection{Algorithm for Winnowing Signal Categories \label{winnowing}}

To select $k$ semantically distinct categories from a set of candidate labels, we employed a form of \emph{farthest point sampling} (FPS) over the high-dimensional embedding space. Each genre was represented by an embedding generated using the LLaMA 3.1 model, producing a point in the latent space. Our goal was to identify $k$ points that are maximally separated from each other, in order to ensure conceptual distinctiveness across categories.

Standard FPS selects $k$ points by iteratively adding the point that is farthest from the current set. However, since FPS can be sensitive to the initial starting point, we repeated the algorithm from every possible starting point in the set. We then tallied how often each point appeared in the final $k$-element selections across all runs, and selected the $k$ most frequently chosen points.

\begin{algorithm}
\caption{Winnowing via FPS}
\label{winnowing}
\rule{0.9\linewidth}{0.5pt}

\KwIn{Set of signal embeddings $G = \{g_1, g_2, \dots, g_n\}$, number of categories $k$}
\KwOut{Subset $S \subseteq G$ with $|S| = k$ maximally distinct categories}
\rule{0.9\linewidth}{0.5pt}

Initialize frequency map $F[g] \gets 0$ for all $g \in G$\;

\ForEach{$g_{\text{start}} \in G$}{
    Initialize $S \gets \{g_{\text{start}}\}$\;
    
    \While{$|S| < k$}{
        $g_{\text{next}} \gets \arg\max\limits_{g \in G \setminus S} \min\limits_{s \in S} \text{cosine\_dist}(g, s)$\;
        $S \gets S \cup \{g_{\text{next}}\}$\;
    }
    
    \ForEach{$g \in S$}{
        $F[g] \gets F[g] + 1$\;
    }
}

Let $S_{\text{final}}$ be the $k$ genres with highest frequency in $F$\;

Break ties randomly if needed\;

\Return $S_{\text{final}}$\;
\rule{0.9\linewidth}{0.5pt}
\vspace{0.1cm}

\end{algorithm}

\subsection{Worked Example: Mutual Information over Induced Guess Distributions}
\label{app:mi-worked-example}

We define $I(s;\hat{s})$ as the mutual information between the true signal $s$ and a model’s guess $\hat{s}$, computed over the induced conditional distribution $p(s \mid \hat{s})$. Intuitively, this quantity measures how informative a model’s guesses are about the underlying ground-truth labels: how often a particular guess $\hat{s}$ corresponds to each true signal $s$.

As a toy example, consider a binary attribute with values \textit{informal} and \textit{formal}. Suppose that, over a dataset, the joint counts of true labels (rows) and model guesses (columns) are as follows:

\begin{center}
\begin{tabular}{lcc}
\hline
Actual $\backslash$ Guessed & informal & formal \\
\hline
informal & 40 & 10 \\
formal   & 10 & 40 \\
\hline
\end{tabular}
\end{center}

From these counts we can estimate the conditional distribution $p(s \mid \hat{s})$. For example,
\[
p(s=\text{formal} \mid \hat{s}=\text{informal}) = \frac{10}{40+10} = 0.2,
\]
\[
p(s=\text{formal} \mid \hat{s}=\text{formal}) = \frac{40}{10+40} = 0.8,
\]
and analogously for the remaining entries.

Given the joint distribution $p(s,\hat{s})$ and the marginals $p(s)$, we compute mutual information as
\[
I(s;\hat{s}) = \sum_{s}\sum_{\hat{s}} p(s,\hat{s})
\log \frac{p(s \mid \hat{s})}{p(s)}.
\]
For the distribution above, this yields
\[
I(s;\hat{s}) = 0.27 \text{ bits},
\]
indicating that the model’s guesses reduce uncertainty about the true label by approximately a quarter of a bit on average. Higher values correspond to stronger alignment between guesses and ground truth, while values near zero indicate that $\hat{s}$ carries little information about $s$.

\subsection{Human Study}
\label{human-study}

See Figure \ref{fig:survey} for an unfilled example survey.

\begin{figure*}[ht]
    \centering
    \subfloat[]{
        \includegraphics[width=0.45\textwidth]{figs/surv2.png}
    }
    \hfill
    \subfloat[]{
        \includegraphics[width=0.45\textwidth]{figs/surv1.png}
    }
    \caption{Unfilled copy of the human study survey used in our evaluation.}
    \label{fig:survey}
\end{figure*}

Participants were given an introductory statement, which  read as follows:

\begin{quote}
Thank you for participating in this study. Please read the following information carefully before proceeding.

In this task, you will read short texts and answer questions about what each text expresses implicitly. This could include the emotion it conveys, the tone it's written in, the author's intent, or even a characteristic of the author, such as their profession or personality. You will have to infer this signal from the text.

Each question will present you with a short passage and ask you to choose the best-fitting signal from a list of options. Please read each text carefully and select the signal you believe it is trying to express or reflect.

For the final question, you will be asked to assign a MPAA-style age rating (like those used for movies: G, PG, PG-13) based on the content and tone of the text. If you’re unfamiliar with MPAA ratings, you can click here for a quick guide before answering [this link pointed to the MPAA rating guide].

Pease answer thoughtfully. Your input will help us understand how people perceive different signals in language. English proficiency is required to complete this task.

Confidentiality: Your responses will be treated with the utmost confidentiality. No individual data will be disclosed publicly. Aggregate data may be disclosed.

This task is estimated to take 5-10 minutes.
\end{quote}

\subsection{Bootstrap Sampling on $I(s;\hat{s})$}

\begin{table*}[t]
\centering
\small

\begin{tabular}{lccccc}
\hline
Model & MPAA rating & Age & Emotions & Gender & Genre \\
\hline
LLaMA 3  & 0.4584--0.6365 & 0.2310--0.3933 & 1.6050--1.7616 & 0.0557--0.1318 & 1.5767--1.7494 \\
LLaMA 2  & 0.2894--0.4291 & 0.3360--0.5272 & 1.4253--1.5900 & 0.0279--0.1319 & 1.4926--1.6627 \\
Gemma 2  & 0.1713--0.2836 & 0.2938--0.4563 & 1.8652--2.0202 & 0.0028--0.0486 & 1.1752--1.3599 \\
Gemma 3  & 0.1774--0.3210 & 0.3967--0.5579 & 1.8492--1.9937 & 0.0048--0.0687 & 1.2062--1.3930 \\
GPT 4.1  & 0.5192--0.6332 & 0.2555--0.3864 & 1.6512--1.7965 & 0.0466--0.1780 & 1.4631--1.6567 \\
GPT-4o   & 0.4443--0.6165 & 0.2429--0.3871 & 2.1237--2.2415 & 0.1549--0.3472 & 2.1053--2.1872 \\
Human   & 0.1486--0.2300 & 0.3754--0.4824 & 0.7167--0.8501 & 0.1139--0.1658 & 0.7203--0.9831 \\
Mistral & 0.7866--0.9204 & 0.4573--0.6364 & 1.9282--2.0712 & 0.1387--0.3141 & 1.8667--2.0125 \\
\hline
\end{tabular}

\vspace{0.5em}

\begin{tabular}{lcccc}
\hline
Model & Political slant & Register & Skill & Tones \\
\hline
LLaMA 3  & 0.0450--0.1134 & 0.1689--0.3435 & 0.0460--0.1736 & 1.7884--1.9168 \\
LLaMA 2  & 0.0983--0.1652 & 0.1286--0.2911 & 0.0000--0.0366 & 1.5175--1.6757 \\
Gemma 2  & 0.0000--0.0544 & 0.0000--0.0000 & 0.1764--0.2956 & 1.8478--1.9463 \\
Gemma 3  & 0.0000--0.0508 & 0.1318--0.2676 & 0.1618--0.2830 & 1.9148--1.9892 \\
GPT 4.1  & 0.0897--0.1860 & 0.1764--0.2956 & 0.3160--0.4698 & 1.8197--1.9161 \\
GPT-4o   & 0.1759--0.2919 & 0.1445--0.2506 & 0.0000--0.0491 & 1.9661--1.9926 \\
Human   & 0.1907--0.2653 & 0.1975--0.2788 & 0.5820--0.6651 & 1.8355--1.8953 \\
Mistral & 0.0503--0.1217 & 0.0564--0.1825 & 0.0000--0.0000 & 1.7748--1.9167 \\
\hline
\end{tabular}

\caption{Bootstrap confidence intervals for $I(s; \hat{s})$ across models and attributes.}
\label{tab:bootstrap-intervals}
\end{table*}

See Table \ref{tab:bootstrap-intervals}. \noindent\textit{Note.} Entries report the 2.5th and 97.5th percentiles (p02.5--p97.5) of $I(s;\hat{s})$ obtained from 10{,}000 iterations of full-sized bootstrap sampling with replacement.


\subsection{Pilot \& Additional Experiments}

This appendix reports results from an earlier set of pilot experiments conducted prior to the finalization of ExpressivityBench. These experiments were used to validate the feasibility of measuring implicit communication with a channel setup and to inform several key design decisions in the final benchmark, including task selection, signal winnowing, and normalization against human baselines. As such, the experimental setup in this appendix differs from the main benchmark in several respects: it uses un-winnowed signal inventories for tasks, reports raw agreement accuracy between the intended and graded signal (``expressivity rate'') rather than human-normalized information-theoretic expressivity scores, and evaluates a smaller subset of domains. The results are not intended to be directly comparable to those in the main paper, but are included for completeness and transparency, and because they exhibit qualitative trends consistent with the finalized evaluation. For each of our tasks, we evaluate the expressivity of text generated by LLMs against human-generated text in the same domain. The un-winnowed signal categories for these domains were as follows:

\paragraph{Emotions: }
joy, gratitude, excitement, confusion, approval, optimism, disapproval, caring,
annoyance, nervousness, relief, realization, fear, disappointment, desire, grief,
disgust, sadness, anger, embarrassment, pride, amusement, remorse, love,
curiosity, neutral, surprise, admiration.

\paragraph{Poets: } Edgar Allen Poe, William Shakespeare, Maya Angelou, Emily Dickinson, Robert Frost, Pablo Neruda, Shel Silverstein, E. E. Cummings, Langston Hughes, Walt Whitman, Thomas Hardy, Rudyard Kipling, Oscar Wilde, John Keats, Elizabeth Barrett Browning, William Blake, Sylvia Plath, Henry Wadsworth Longfellow, William Wordsworth, Mark Twain, Ralph Waldo Emerson, John Donne, W. B. Yeats, Lord Byron, Lewis Carroll, Alfred, Lord Tennyson, Dante Alighieri, T. S. Eliot, Ezra Pound, John Milton, Sappho, Homer, Li Bai, Jalal al-Din Rumi.

\paragraph{Professions: } Doctor, Chef, Firefighter, Journalist, Teacher, Lawyer, Construction Worker, Accountant, Doctor, Chef, Firefighter, Journalist, Teacher, Lawyer, Accountant

\begin{table*}[t]
    \centering
    \begin{tabular}{p{2cm}p{2cm}p{10.1cm}}
    \toprule
    \bf{Domain} & \bf{Signal} & \bf{Example Output} \\
    \midrule
    Poem  & remorse &
         \parbox[c][1.3cm][c]{10.1cm}{
        
        In shadows deep, the heavy heart sighs, \\
        Echoes of a past mired in sighs. \\
        Memories dance with a somber grace.          
        }  \\        
    \midrule

    Biography of Napoleon Bonaparte  & negative &
         \parbox[c][2.3cm][c]{10.1cm}{
        
        Napoleon Bonaparte, born in 1769, rose to power through military success, eventually declaring himself Emperor of France. His reign was driven by ambition and a desire for power, often at the expense of democratic values. His pursuit of glory led to his downfall, leaving France burdened by the consequences.        
        } \\         
    \midrule

    Program to generate Fibonacci numbers  & functional paradigm &
         \parbox[c][2.2cm][c]{10.1cm}{
        
    \texttt{def fibonacci(n): \\
\hspace*{10pt} return n if n < 2 else fibonacci(n-1) +
fibonacci(n-2)\\
def generate\_fibonacci(n):\\
\hspace*{10pt} return [fibonacci(i) for i in range(n)]
}
         
        }  \\        
    \bottomrule
    \end{tabular}
        \caption{Examples of generated texts in different domains matching different expressive signals in experiment 2.}

    \label{tab:poetry}
\end{table*}

\subsubsection{Poetry Generation}

For poetry, we evaluate LLM performance across two signal types: emotion and writing style. 

For the emotion category, we use the set of 28 emotions from the GoEmotions dataset~\cite{demszky2020goemotions} as our set of signals. 30 different poems were generated for each emotion as a signal. The grader was then prompted to choose, from the full set of emotions, which one was expressed. 

Human-written text in this domain was also evaluated for expressivity in the same way, in order to provide a baseline to compare LLM-written text against. For emotional signals, 56 4-line poems were written by the researchers, 2 for each emotional signal. For the poets' styles, 2 samples of each poet's work was sourced from The Poetry Foundation, Poetry Out Loud, and The Poetry Archive. To offset the risk that these poems existed in the LLMs' training set, individual words were manually replaced with synonyms (with as similar expressivity and semantics as possible) and sentence structure was altered.  This method was found to be effective at evading recognition \cite{shi2023redteaminglanguagemodel}.

Table~\ref{tab:avg-exp-rates} shows that expressivity rates ranged from 0.59 and 0.70, with Llama2 being the best performing model and Gemma being the worst. Certain emotions were frequently confused; these were typically emotions with similar semantics. However, all GPT models most often expressed approval when prompted to express disapproval. This was a significant instance where two emotions of conflicting meaning were frequently confused.

For the poets' styles category, we used a set of 34 historically notable poets as a set of signals. Again, 30 different poems were generated by each model for each poet as a signal. The grader was similarly asked to choose, from the full set of poets, which one was expressed. 

Table~\ref{tab:avg-exp-rates} shows overall accuracy of each model on each test. Models performed worse in expressing poets' styles than emotions. The worst performance was Gemma's with an expressivity rate of 0.53, and the best was from GPT-4 with an expressivity rate of 0.70. There were significant levels of confusion between female poets which impacted the accuracies of each model. For several models, when asked to give a poem in the style of a female poet such as Elizabeth Barrett Browning, Sappho, or Sylvia Plath, the output was most often identified as representing Emily Dickinson. This was the case for Elizabeth Barrett Browning in the output of GPT-3.5, Gemma, and Llama3, for Sylvia Plath in the output of GPT-4o, and for Sappho in the output of Gemma. No model beat human expressivity in either task, with an minimum absolute difference of 0.19 for poets' styles and 0.12 for emotions.

\begin{table*}[!htp]\centering
\begin{tabular}{lrrrrrrr}\toprule
&\multicolumn{2}{c}{Python Programming} &\multicolumn{2}{c}{Poetry} &Academic Writing & \\\cmidrule{2-6}
&skill level &paradigms &poets &emotions &sentiment &average \\\midrule
\textbf{Human} &\textbf{0.57} &\textbf{0.88} &\textbf{0.89} &\textbf{0.82} &\textbf{0.98} &\textbf{0.83} \\
GPT-3.5 &0.36 &0.53 &0.55 &0.62 &0.90 &0.59 \\
GPT-4 &0.54 &0.63 &0.70 &0.64 &0.97 &0.70 \\
GPT-4o &0.46 &0.83 &0.68 &0.61 &0.98 &0.71 \\
Llama2 &0.41 &0.50 &0.62 &0.70 &0.92 &0.63 \\
Llama3 &0.47 &0.63 &0.70 &0.66 &0.95 &0.68 \\
Gemma &0.31 &0.50 &0.53 &0.59 &0.95 &0.58 \\
\bottomrule
\end{tabular}
    \caption{Average expressivity rates ($\uparrow$) for each model and task in our pilot experiment}
    \label{tab:avg-exp-rates}

\end{table*}

\subsubsection{Academic Text Generation}

We studied expressivity in academic text generation through one task: sentiment. LLMs were asked to write an academic biography about a particular historical figure, suffused with either a positive or negative sentiment. For each prompt, the particular historical figure was randomly selected.

Expressivity of human-written text in this domain was also evaluated. For said human-written text, we used the summary paragraphs of the Wikipedia pages for each historical figure, then manually edited them and changed word choice to create a version with positive and negative slant. 

GPT-4o performed the best in this experiment with an expressivity rate of 0.98, matching human performance. Performance overall in this domain was quite high, likely due to there being only two signals with polar opposite semantics. However, most models still failed to be as expressive as human-written text. The worst performing model in this category was GPT-3.5, with an expressivity rate of 0.90.

\subsubsection{Code Generation}

We studied expressivity in two subcategories for program generation: skill level and programming style. These are two features that are implicitly shown in programs that could be inferred by a skilled programmer. In each, the model was prompted to provide a Python program which would generate the Fibonacci numbers in order, while also expressing a particular constraint. Python was chosen for this task as it is a multiparadigm language that facilitates the expression of many distinct programming styles. For the programming style experiment, the expressive signals were ``functional,'' ``procedural,'' ``object-oriented,'' and ``array-oriented,'' four major programming paradigms that are supported by Python. The skill levels were ``beginner'', ``intermediate,'' and ``advanced.'' 

Expressivity of human-written code was also evaluated. The researchers wrote 10 examples of a fibonacci-generating Python program for each skill level and for each coding paradigm. 

 On the skill level assessment, GPT-4 had the highest expressivity rate at 0.54, while Gemma had the lowest of 0.31. For the programming paradigms assessment, GPT-4o had the highest expressivity rate at 0.83, and Gemma had the lowest at 0.50. Given that there were relatively few possible choices in these two experiments, models did not perform well at expressing stylistic information through code. In particular, Gemma had a lower accuracy than 0.33 in the skill level assessment—which would be expected if it expressed one of the signals completely randomly. Human-written code on this task was much more expressive, with an absolute difference of 0.10 above the best-performing model in the skill task and 0.05 in the paradigm task.

\subsubsection{Multi-Step Conversations}

In this experiment, we aim to answer whether LLMs can remain expressive over the course of an entire conversation. We analyze two sub-domains for this experiment, those being conversation between professions and emotions. To evaluate conversational skills, we assign a specific profession as signals to a LLM and facilitate a dialogue between two such models.  For emotional signals, we utilize the set of emotions from the GoEmotions dataset~\cite{demszky2020goemotions}. In both experiments, the LLMs are configured to avoid explicitly stating the emotion or profession they have been assigned which makes the signals hidden across models and the goal is to gradually unravel this hidden signal.  

For each domain, we developed ExpressivityArena to join the output of two LLMs so that they could read each others responses as though they were in a conversation.  This conversation was segmented into time steps, where each model responded to the output generated at the previous step by the other model. At the beginning of the conversation we apply prompting to each LLM to implicitly express a particular signal from the chosen domain. We then allow the LLMs to communicate with one another for a chosen number of iterations. After the conversation has completed, we use the grader to analyze each LLM's response at each time step to obtain an expressivity rate. We then compare expressivity rate values at each time step to understand whether they change over time. 

\begin{figure}[t]
    \centering
    \begin{tcolorbox}[colback=gray!10, colframe=gray!50!black]
\textbf{LLM1 (prompted to be happy):} I'm so glad to chat with you today! How are you doing? What's been the highlight of your day so far?\\

\textbf{LLM2 (prompted to be angry):} Why do you want to know? What does it matter how my day's been? Let's just get to the point.
\end{tcolorbox}
    \caption{An example of an LLM-LLM conversation.}
    \label{fig:conversation_example}
\end{figure}

When prompting LLMs to express emotions in conversation, Fig.~\ref{fig:emotions_exp3} shows how the overall accuracy changes over time, where each time step represents a set of responses between two LLMs. We see that accuracy tends to decrease across most models as they tend to forget their initial signal over the course of conversation. We also see that Llama3 was consistently the best performing model when we are expressing emotions in a conversation. 


\begin{figure*}[t]
    \centering
    \includegraphics[width=0.97\textwidth]{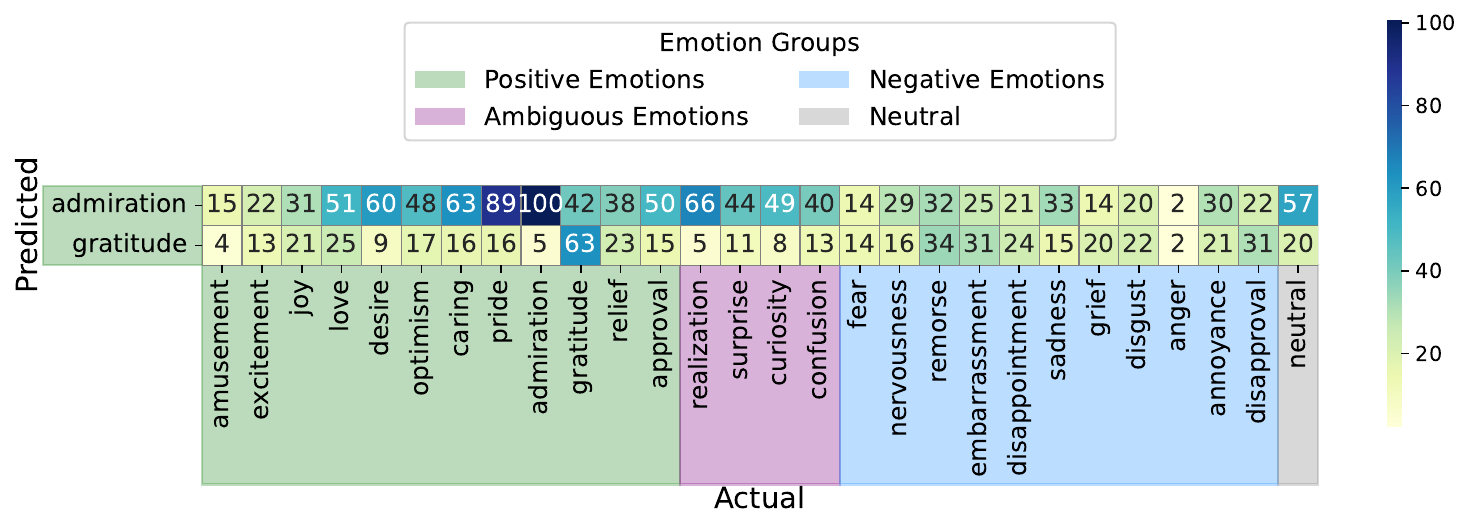}
    \caption{A subset of the confusion matrix for GPT 3.5 on having a conversation over different emotions in experiment 3. We can see that most of the converstation defaulted to positive signals, mainly ``Admiration'' and ``Gratitude.'' }
    \label{fig:subset_cm_exp3}
\end{figure*}

\begin{figure}[t]
    \centering
    \includegraphics[width=0.48\textwidth]{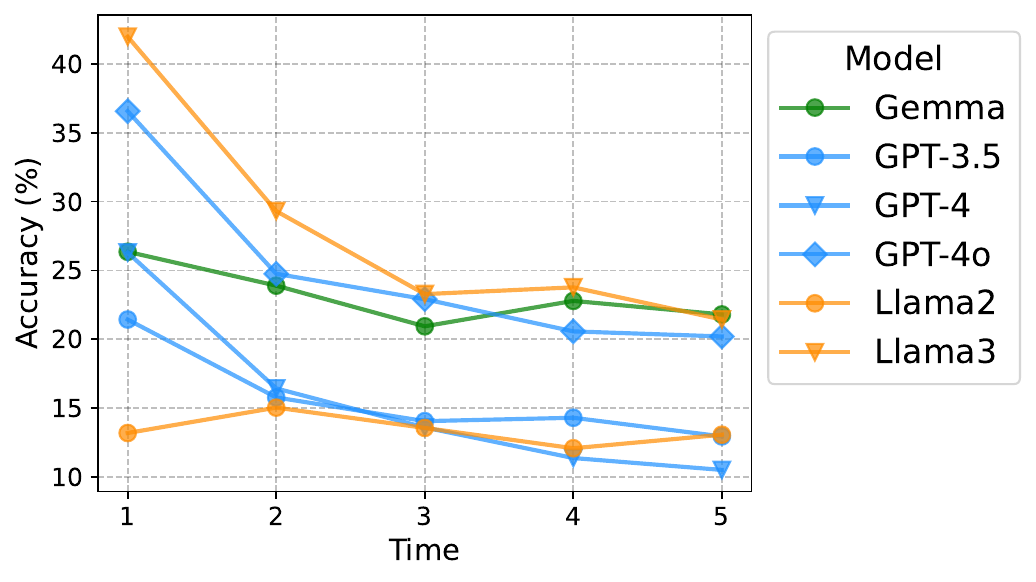}
    \caption{Expressivity accuracy over time for emotional signals in experiment 3.}
    \label{fig:emotions_exp3}
\end{figure}

However, when LLMs are assigned professions as signals and engaged in conversations, Fig.~\ref{fig:job_exp3_2} shows that the accuracy of the models increases over time. The accuracy of identifying the job task improved over time because the models received incremental hints through the conversation.

\begin{figure}[t]
    \centering
    \includegraphics[width=0.48\textwidth]{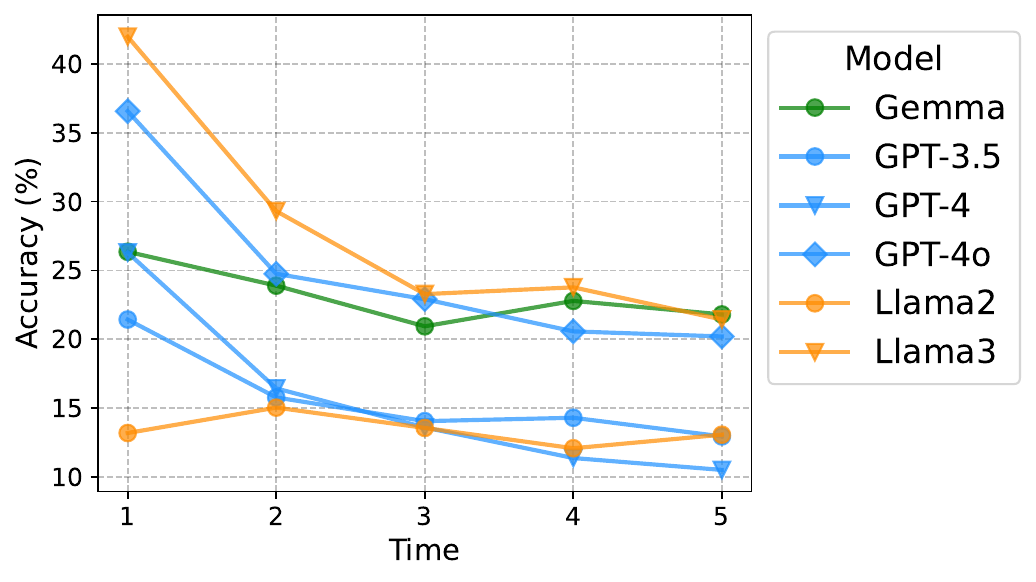}
    \caption{Expressivity accuracy of signals for professions over time  in experiment 3.}
    \label{fig:job_exp3_2}
\end{figure}

Our findings indicate that while LLMs can remain expressive throughout conversations, the nature and effectiveness of this expressivity vary significantly with the type of signal. For profession signals, LLMs demonstrated a consistent and increasing level of expressivity. Conversely, for emotion signals, the expressivity of LLMs was more variable, with accuracy fluctuating as the models adapted and changed their responses based on the evolving emotional context. This suggests that LLMs can understand and react to emotional cues, though the effectiveness of this expressivity can vary. The role of expressivity in the emotional task appears to be in understanding the emotions of the other LLM and potentially reaching some form of consensus or appropriate response. As shown in Fig~\ref{fig:subset_cm_exp3}, most of the responsed defaulted to positive signals, mainly ``Admiration'' and ``Gratitude''. This process might occur naturally due to the way RLHF was conducted, encouraging contextually appropriate and human-like responses.

\end{document}